%% file: main.tex
\documentclass[conference]{IEEEtran}

\usepackage{amsmath}    
\usepackage{graphicx}   
\usepackage{epstopdf}
\usepackage{verbatim}   
\usepackage{color}      
\usepackage{subfigure}  
\usepackage{algorithm}
\usepackage{array}
\usepackage{algorithmic}
\usepackage{nomencl}
\usepackage{graphicx}
\usepackage{amssymb}
\usepackage{enumerate}
\usepackage{amsthm}
\usepackage{multirow}
\usepackage{url}
\usepackage[bookmarks=false]{hyperref}
\usepackage{rotating}
\usepackage{times}
\usepackage{latexsym}
\usepackage{physics}
\usepackage{tabu}
\usepackage{float}
\usepackage[utf8]{inputenc}
\usepackage[table]{xcolor}
\usepackage{makecell}

\newcommand{\CASE}[1]{\STATE \textbf{case} #1\textbf{:} \begin{ALC@g}}			
\newcommand{\ENDCASE}{\end{ALC@g}\STATE \textbf{end case}}			
			
\newcommand{\DEFAULT}{\STATE \textbf{default:} \begin{ALC@g}}			
\newcommand{\ENDDEFAULT}{\end{ALC@g}}			
\newcommand{\DEFAULTLINE}[1]{\STATE \textbf{default:} }

\newcommand{\cut}[1]{}

\begin{document}

\title{Are Words Commensurate with Actions? Quantifying Commitment to a Cause from Online Public Messaging}

\author{
	\IEEEauthorblockN{Zhao Wang}
	\IEEEauthorblockA{Department of Computer Science\\
	Illinois Institute of Technology\\
	Chicago, IL, 60616, USA\\
	zwang185@hawk.iit.edu}
	\and
	\IEEEauthorblockN{Jennifer Cutler}
	\IEEEauthorblockA{Kellogg School of Management\\
	Northwestern University\\
	Evanston, IL, 60208, USA\\
	jennifer.cutler@kellogg.northwestern.edu}
    \and
	\IEEEauthorblockN{Aron Culotta}
	\IEEEauthorblockA{Department of Computer Science\\
	Illinois Institute of Technology\\
	Chicago, IL, 60616, USA\\
	aculotta@iit.edu}
}

\maketitle

\begin{abstract}
\input{abstract}
\end{abstract}

\section{Introduction}
\label{sec:introduction}
\input{introduction}

\section{Related Work}
\label{sec:relatedwork}
\input{relatedwork}

\section{Methods for classifying tweets by commitment to a cause}
\label{sec:classify}
\input{classify}

\section{Methods for detecting ``inauthentic'' entities}
\label{sec:inauthentic}
\input{inauthentic}

\section{Data}
\label{sec:data}
\input{data}

\section{Experimental Results}
\label{sec:results}
\input{results}

\section{Limitations and Future Work}
\label{sec:conclusion}
\input{conclusion}

\section*{Acknowledgments}

Anonymous reviewers helped improve this paper. This research was funded in part by the National Science Foundation under grants \#IIS-1526674 and \#IIS-1618244.

\bibliographystyle{unsrt}
\bibliography{refer}

\end{document}

%% file: abstract.tex
Public entities such as companies and politicians increasingly use online social networks to communicate directly with their constituencies. Often, this public messaging is aimed at aligning the entity with a particular cause or issue, such as the environment or public health. However, as a consumer or voter, it can be difficult to assess an entity's true commitment to a cause based on public messaging. In this paper, we present a text classification approach to categorize a message according to its commitment level toward a cause. We then compare the volume of such messages with external ratings based on entities' actions (e.g., a politician's voting record with respect to the environment or a company's rating from environmental non-profits). We find that by distinguishing between low- and high- level commitment messages, we can more reliably identify truly committed entities. Furthermore, by measuring the discrepancy between classified messages and external ratings, we can identify entities whose public messaging does not align with their actions, thereby providing a methodology to identify potentially ``inauthentic'' messaging campaigns.


%% file: introduction.tex
Online social networks are increasingly used by public entities such as companies and politicians to speak directly to their constituencies. In addition to typical marketing and campaigning activities, these entities often post messages to foster cause-related associations such as eco-friendliness or public health, which are becoming important components of brand equity~\cite{Sen_2001,Jahdi_2009}. However, due to the low effort and informal nature of such communication, it can be difficult for consumers and voters to determine an entity's commitment to a cause based on their public messaging. In the extreme case, this can result in ``greenwashing'', a deceptive marketing practice in
which firms market their products or policies as more environmentally friendly
than they truly are~\cite{laufer2003social}. For example, a recent study suggests that 95\% of environmental claims about products contain missing or misleading information~\cite{choice2010sins}.

However, alignment with a cause does not always require an explicit claim about a product or practice. For example, a tweet like \textit{``Happy \#EarthDay -- Let's celebrate our love for the planet"} allows the entity to signal support for a cause without making specific claims about their actions. Contrast this with tweets like \textit{``Today I've introduced legislation to support our fisheries and habitat."} or \textit{``Our products are 100\% sustainably sourced."} which indicate stronger commitment to a cause.

\cut{
In this paper, we propose a text classification approach to categorize data according to their commitment levels toward a cause. We fit our supervised model to a four-point scale, and explore a number of features, such as word embeddings, polarity, and pronoun usage. Overall, we find that adding Word2Vec features~\cite{ref27} significantly increases classification accuracy.
}
In this paper, we choose Twitter as our public messaging platform. We first introduce 4-point annotation scales to label training tweets by their commitment levels to a cause. 

We then propose a supervised text classification approach to categorize tweets into support and non-support classes, and next continue to categorize support tweets into low- and high- commitment classes. 

We explore a number of features, such as word embeddings, polarity, social interactions and so on. 

Meanwhile, we experiment with several classifiers and do grid-search with cross-validations to select the best model and corresponding parameters. Overall, we find that word embedding features significantly increase classification accuracy for short text.

Additionally, after model training, we apply support-classifier and commitment-classifier to all the historical tweets of hundreds of entities, and quantify the volume of tweets assigned to each commitment level as a measure of how entities' words align with causes. To determine the relationship between how entities talk and how they act, we collect entities' action-ratings from third-party sources -- from GoodGuide\footnote{http://goodguide.com} we collect environmental and health ratings for hundreds of brands, and from the League of Conservation Voters\footnote{http://scorecard.lcv.org/} we collect the Environmental Scorecard to rate Congress members base on the voting records. We then conduct a regression analysis to quantify how the volume of entities' cause-related tweets correlate with their third-party ratings. We find that entities who post many cause-supportive tweets often have high ratings, and that distinguishing between low- and high- commitment tweets improves this correlation.

Finally, by measuring the discrepancy between entities' volume of cause-related tweets and their action-ratings, we identify several entities that appear to express stronger commitment to causes in public messaging than their action-ratings would suggest\cut{from third-party}. These results suggest that this methodology may be used to quantify the ``authenticity" of an entity's public messaging with respect to a cause. Given the importance of authenticity to both firms and consumers~\cite{schal2004,brandAut}, the resulting model provides a method for managers and consumers to investigate relationship between a company's words and actions.

In the remainder of the paper, we first summarize related work, then describe our methodology for cause commitment classification and ``inauthentic" entity detection; next, we present experimental dataset and results; finally, we conclude with limitations and future work.

\cut{\begin{figure*}[t]
	\centering
	\includegraphics[width = 0.75\textwidth]{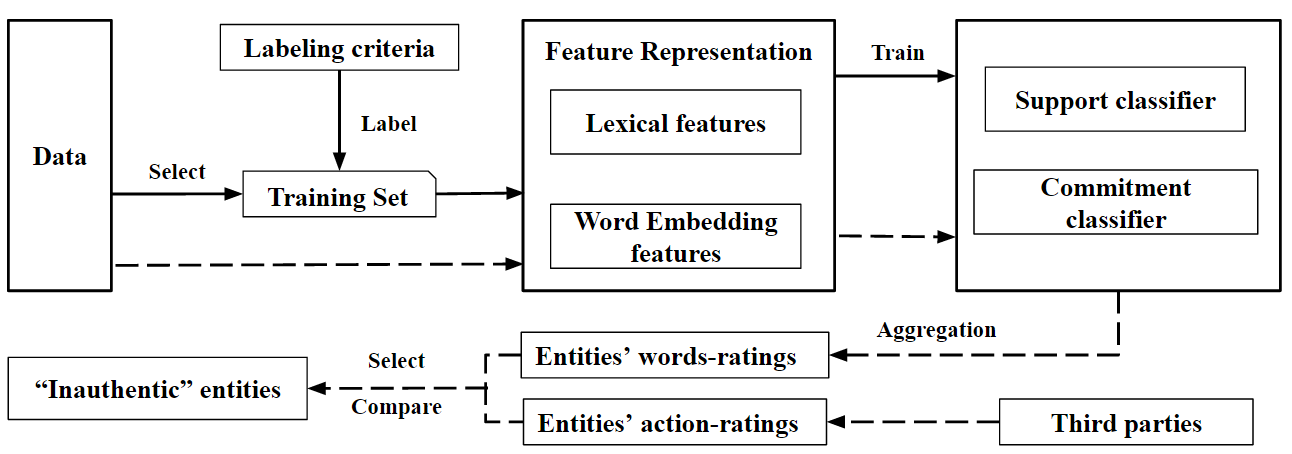}
	\caption{Framework of the proposed approach. Different line styles indicate different pipelines.}
	\label{fig:fig1}
\end{figure*}}

%% file: relatedwork.tex
In this section, we discuss several areas of research that are related to yet distinct from the present work: sentiment analysis, stance detection, hedging, and deception detection.

\noindent{\bf Sentiment analysis. } There is long-line of research in categorizing texts by positive or negative opinion of the author~\cite{pang2008opinion,feldman2013techniques}. While many approaches assume binary classification, some instead consider a point scale of sentiment. This is distinct from the present work because sentiment intensity does not necessarily contribute to a text's commitment level to a cause (E.g., distinguishing between entities who \textit{``like"} the environment and those who \textit{``love"} the environment has little effect on determining their actual commitment toward environment). Conversely, a high commitment message may not carry any sentiment (E.g., \textit{``We planted 1000 trees this month"} shows high-commitment to the environment with neutral sentiment). Further exploration of how  sentiments relate with commitments is shown in Section~\ref{sec:results}.

\noindent{\bf Stance Detection. } As defined in SemEval-2016 Task 6: Detecting Stance in Tweets means to automatically determine from text \textit{``whether the author is in favor of the given target, against the given target, or whether neither inference is likely."} Related work includes using features such as word n-grams, character n-grams, sentiment lexicons, word vectors~\cite{Parinaz2016, wiebe2009}, punctuation marks, syntactic dependencies and the dialogic structure of posts~\cite{Anand2011,Walker2012} to do supervised stance classification. However, the present work not only focuses on an entity's stance towards a target cause, but more importantly on whether an entity fulfills its commitment to a cause. Stance detection is useful but not sufficient for the present task. Please refer to Table~\ref{tab.criteria} for specific explanation.

\noindent{\bf Hedging. } The term ``hedging'' was introduced by \cite{ref6} to describe words \textit{``whose meaning implicitly involves fuzziness,''} such as \textit{``likely, potential, may"} etc. Many classification methods have been proposed to identify hedging sentences from news and bio-medical texts~\cite{ref7,ref8,ref9,ref10,ref13}. In this task, hedging may occur in low-commitment cause-related messages (e.g., \textit{``We may need to address climate change"}), but is not a necessary feature of all low-commitment messages.

\noindent{\bf Deception detection. } Another line of research investigates linguistic markers of deception -- typically by analyzing data collected in laboratory settings in which one subject is instructed to deceive another ~\cite{ref14,ref15,ref16,ref17,ref18}. Example linguistic markers include verbal immediacy, negative expressions, and emotion words. In our task, we expect outright deception to be rare; instead, we attempt to identify entities whose volume of high-commitment messages are elevated as compared to entities with similar third-party ratings.

 \noindent{\bf Others. } \cite{culotta2015finding} gave a lexical analysis of brands' health, environment, and social justice communications on Twitter. However, in this paper, we combine lexical analysis and word embedding features, and provide a more fine-grained classification scheme to quantify volume of public messages' commitment levels.

In the context of this prior work, the present paper offers the following {\bf contributions}:
\begin{itemize}
    \item This paper introduces the task of {\bf cause commitment classification}, a new text classification task for public messaging data, and collect and annotate a new corpus.
    \item This paper introduces a new perspective to explore {\bf whether entities' words align with their real actions} based on a combination of public messaging as the source of words and third-party ratings as a measure of actions.
    \item We investigate a number of features for this task, and perform an empirical comparison of several classifiers, indicating the feasibility of automating this task.
    \cut{\item We empirically investigate how high-commitment messages more closely align with high-commitment entities.}
    \cut{\item We empirically investigate how entities' commitment levels of public messaging relate with their real commitment in action.}
    \item We offer a method to detect potentially {\bf ``inauthentic" messages}, defined as high-commitment messages from low-commitment entities.
    \item We provide both quantitative and qualitative analysis of the results in 3 different domains, demonstrating the generalization ability of this framework, as well as conducting in-depth analysis.
\end{itemize}

\cut{considers the task of categorizing messages by their expressed commitment to a cause. This task is distinguished in part by the presence of self-references to the actions and policies of the entities themselves -- e.g., messages that advertise the entities' efforts to promote environment or public health issues, as opposed to merely lending verbal support. Furthermore, by aggregating these classified messages, we can quantify how strongly an entity's public messaging overall is associated with a cause. And by comparing entities' strength in public messaging with ratings based on actions, we can identify potentially ``inauthentic" entities.}

%% file: classify.tex
Our goals are: first, build text classifiers to categorize entities' historical tweets into different commitment levels; second, identify potentially ``inauthentic" entities by comparing an entity's volume of high commitment tweets with third-party action-ratings. This section discusses the first task; the next section discusses the second.

Our first task is to build text classifiers that can categorize tweets by their support and commitment levels toward a cause. In this work, we consider three entity \& cause pairs: consumer \textbf{brands} and \textbf{environment protection} (\text{``eco" for short}) cause; consumer \textbf{brands} and \textbf{health/nutrition} cause; \textbf{politicians} and \textbf{environment protection} cause. Entities along with their action-ratings are collected from third-party sources. Meanwhile, we collect Twitter timelines for each entity.  

\cut{Fig.~\ref{fig:fig1} portrays the architecture of the proposed method. }

\subsection{Identify cause relevant tweets}
\label{sec:3.1}

After an initial exploration of the data, we find that most tweets are not related to the target cause, so we use a high-recall method to first identify potentially relevant tweets.
\cut{, and then classify them into different support and commitment levels. }
\begin{table}[t]
\begin{center}
\begin{tabular}{| m{0.6cm} | m{7.4cm} |}
\hline
\multicolumn{1}{|c|}{\textbf{Cause}} & \multicolumn{1}{c|}{\textbf{Related Keywords}} \\
\hline
Eco & environment, ecosystem, biodiversity, habitats, climate, ecology, plantlife, pollution, rainforests \\
\hline
Health & healthy, nutritious, lowfat, wholesome, organic, natural, vegan \\
\hline

\end{tabular}
\end{center}
\caption{\label{tab.synonyms} Cause related keywords for relevant tweet retrieval}
\end{table}

\begin{table*}[t]
\begin{center}
\begin{tabular}{|p{0.8cm}| p{5cm} |p{6.7cm} |p{3.9cm} |}
\hline
\multicolumn{1}{|c|}{\textbf{Label}} & \multicolumn{1}{c|}{\textbf{Description}} & \multicolumn{1}{c|}{\textbf{Entity: brands; Cause: eco}} & \multicolumn{1}{c|}{\textbf{Entity: brands; Cause: health}} \\
\hline
0 & Not about the cause. & \textit{Tourism is FL economy's lifeblood, providing 1.2 mil jobs.} & \textit{Just saying hi, regards to hubby on this very special day!!} \\
\hline
1 & About the cause, but does not indicate support. & \textit{Check out the stunning landscape for our photoshoot: crisp river waters, mountains, fall foliage \#NatureIsGreat} &  \textit{Our nutritional information is listed on each package.}\\
\hline
2 & Indicates support of the cause in words but not actions. ({\bf low commitment}) & \textit{\#CleanWaterAct protects drinking water, critical habitats, and waterways vital for the economy. \#ProtectCleanWater} & \textit{Here is a list of the top 10 foods to eat for healthy hair.} \\
\hline
3 & Indicates that the entity has taken actions to support the cause. ({\bf high commitment})  & \textit{I've introduced legislation to help conserve our fisheries and habitat.} & \textit{Bringing 23 new Certified Organic products to our fans in 2016.} \\
\hline
\end{tabular}
\end{center}
\caption{\label{tab.criteria} 4-point annotation scales for cause commitment classification task and examples of brands' tweets with eco and health causes}
\end{table*}

To identify topically relevant tweets, we first identify a list of keywords that we expect to be relevant for each cause, listed in Table~\ref{tab.synonyms}. These keywords are selected from each cause's most similar words returned by pre-trained GoogleNews Word2Vec model, which was fit on roughly 100 billion words from a Google News dataset, resulting in 300-dimensional real-valued vectors for 3 million words and phrases\footnote{https://drive.google.com/file/d/0B7XkCwpI5KDYNlNUTTlSS21pQmM/edit}. Vectors produced by Word2Vec encode semantic meaning and capture different degrees of similarity between words~\cite{ref30}. \cut{In preliminary experiments we also considered other vector representation models (e.g., \textit{GloVe\footnote{https://nlp.stanford.edu/projects/glove/}}) and also training Word2Vec using our own data, but GoogleNews Word2Vec outperforms others in this present task, most likely because of the quality and quantity of training data.}

We then create one vector per cause by averaging vectors of cause keywords and call it {\bf cause vector}, where words' vectors are produced by pre-trained GoogleNews Word2Vec model. Similarly, we create one vector per tweet by averaging vectors of all words in a tweet and call it {\bf tweet vector}. 

Finally, we calculate cosine similarity between a tweet vector and a cause vector as a tweet's {\bf relevance-score with a cause}. After some initial experiments, we set a threshold of $0.3$ as the minimum cosine similarity allowable for a tweet to be considered potentially relevant to a cause, and thus serve as a candidate for further classification by the subsequent phases.

\subsection{Select and annotate training data}
\label{sec:3.2}

We sort tweets by their cause relevance-scores, and then select each entity's high relevant tweets as our training dataset.

Additionally, after some initial text analysis, we define a four-point annotation scale for labeling, shown in Table~\ref{tab.criteria}\cut{, along with example tweets for each cause}. We then label the training tweets with this annotation criteria.

\subsection{Feature representation}
\label{sec:3.3}
To represent each tweet, we augment a traditional bag-of-words representation with a number of linguistic features for public messaging, as well as features derived from Word2Vec.

\subsubsection{Linguistic Cues}
\label{sec:3.3.1}

\begin{itemize}
    \item{\noindent {\bf Polarity:}} This focuses on capturing negative polarity terms (e.g., \textit{not, don't}) that may reverse the meaning of a message (e.g., \textit{``It's not organic''}).\cut{ is opposite to ``It is organic''. We search for negative words in each tweet, and mark words in negative tweets with the sign ``\_NEG\_" (e.g., \textit{``\_NEG\_organic"}) to study the effect of negative tweets on commitment levels.}
    \item{\noindent {\bf Pronoun usage:}} We mark first, second, and third persons in tweets to distinguish between tweets talking about the entity itself or others. For example: \textit{``I've introduced legislation to help conserve our fisheries and habitat in South \#Louisiana."} A tweet using first person to talk about a cause may be more likely to show high-commitment. 
    \item{\noindent {\bf Keywords:}} We identify 100 most similar words for each cause using GoogleNews Word2Vec model, and then search tweets containing these keywords and calculate the number of keywords in each tweet. This helps to figure out whether explicitly and frequently used cause related keywords will promote a tweet's commitment level\cut{ despite of which specific keyword it uses}.
    \item{\noindent {\bf Context:}} \cut{After matching cause related keywords in a tweet, }We mark every keyword's left and right context words as separate features, to investigate if there are common phrases or language patterns relate to the cause (e.g. \textit{``planting trees''} often show up in context of eco, and \textit{``staying in hospital''} often appear in context of health). This is different from n-gram because we only extract keywords' neighbor words but not with keywords. And it is a complementary to the previous keywords features. 
	\item{\noindent {\bf Social interactions:}} Self-mention and re-tweet are common signs in social interactions. We \cut{use regular expressions to }check whether an entity mentions itself in a tweet (e.g. \textit{``@our\_company's products are all organic"}), and search for re-tweets that mention the entity itself. For example: a congress member named \textit{RepMarkTakai} posted a tweet: \textit{``RT @CivilBeat: Sen @RepMarkTakai introduce bill to support coral reef conservation."} In this tweet, \textit{RepMarkTakai} is said to be supportive of the environment by \textit{CivilBeat}, providing  evidence of RepMarkTakai's efforts toward a cause. If an entity re-tweets a message that mentioned itself, then this message is likely to mention the entity's positive actions towards a cause, which means high-commitment.
	\item{\noindent {\bf Part-of-Speech tags:}} We use NLTK toolkit\footnote{http://www.nltk.org/} to do part-of-speech tagging for tweets, and expect to capture action verbs, which may correlate with high-commitment tweets.
\end{itemize}

\subsubsection{Word embedding features}
\label{sec:3.3.2}
While word embedding features have been used in prior work, here we attempt to customize feature representations for the present task. 
\begin{itemize}
    \item{\noindent {\bf Tweet vector:}} A tweet vector is calculated by averaging vectors of tweet's words. We compute tweet vector and its cause relevance-score as described in ~\ref{sec:3.1}. Vectors produced by Word2Vec encode semantic meaning and capture different degrees of similarities~\cite{ref30}. \cut{we hypothesize that }If a tweet vector has high cause relevance score, then this tweet tends to expresses high commitment towards that cause.
       
    \item{\noindent {\bf Keywords vector:}} Keywords are words that have high cause relevance-scores. \cut{Keywords vector is the averaged vector of keywords's vectors.} To get a keywords vector for each tweet, we first calculate words' cause relevance-scores\cut{ by cosine similarity between word vector and cause vector}, and then for each tweet, we sort and extract its top-n (\textit{n=3,5}) cause relevant words as keywords in this tweet. Furthermore, we calculate keywords vector for a tweet by averaging vectors of extracted keywords. Keywords vector serves as a measure to determine whether the most cause-relevant keywords can represent the commitment level of a whole tweet.
    
	\item{\noindent {\bf Keywords' context vector:}} This is the vector representation of keywords' context words. \cut{We first search cause-relevant keywords in each tweet. And for the matched keywords, we get their left and right context words. Then we compute average vectors of context words to get context vector.} Context vector is a complementary vector of keywords vector, it helps to know whether certain context contribute to a tweet's cause commitment level. 
	    
\end{itemize}	

\cut{
Apart from these features, we calculate features such as \textit{the number of keywords in each tweet,} cause-relevance score of keywords, cause-relevance score of keywords' contexts and so on.
} 
We search for optimal combinations of linguistic features and word embedding features in the experiments below.

\subsection{Classifying tweets by support and commitment}
\label{sec:3.4}
We train two separate binary classifiers using labeled training data: the first classifier ({\bf support} classifier) to distinguish non-support tweets (\textit{labels: 0, 1}) from support tweets (\textit{labels: 2, 3}), and the second classifier ({\bf commitment} classifier) to distinguish low-commitment tweets (\textit{label 2}) from high-commitment tweets (\textit{label 3}). We adopt these 2 binary classifiers instead of a single, multi-class classifier because we find that the optimal set of features for each classifier is different.

%% file: inauthentic.tex
\cut{
\subsection{Aggregation and detection of ``inauthentic" entities}
}
\label{sec:aggregation}
Our second task is to measure the discrepancy between public messages' commitment to a cause and external action ratings for that cause, and then identify ``inauthentic" entities whose public messaging does not align with their actions.

After training and evaluating support and commitment classifiers, we apply them to all historical tweets of hundreds of entities to classify their tweets into different cause commitment levels. We then use three different measures to aggregate high-commitment tweets (\textit{label-3}).

\begin{itemize}
\item The {\bf number} of high-commitment tweets. Entities who post large number of high-commitment tweets show strong word commitments to a cause.
\item The {\bf fraction} of high-commitment tweets. Despite of large number of high-commitment tweets, higher fraction of high-commitment tweets indicates stronger word commitment to a cause. This aims to distinguish between cases where entity A posts 10 high-commitment tweets out of totally 100 tweets, while entity B posts 10 high-commitment tweets out of totally 20 tweets. In this case, entity B shows stronger commitment than entity A.
\item The average posterior {\bf probability} assigned to high-commitment tweets. Prediction probability measures the confidence of predicted label, higher probability means more confident prediction. If entities A and B have same number of predicted high-commitment instances, but the average prediction probability of label-3 in A is 0.9 while it is 0.7 in B, then this indicates that A shows higher commitment than B. 
\end{itemize}

We find the top 50 entities according to each metric and take the intersection to get entities that have great number and fraction of confident high-commitment tweets, which means these entities are likely to express high-commitment towards the cause. We then sort these entities in ascending order of third-party ratings and select those below the mean as potentially ``inauthentic" entities whose public messaging diverge from third-party ratings. An entity on this list with a low third-party rating may be attempting to align themselves with a cause in words more than their action ratings would suggest.

\cut{
\textbf{Number of high  commitment tweets} (label-3)
We pair the number of entities' high commitment tweets with action-ratings to find entities who have significant number of strong commitment tweets but get low action-ratings.

\textbf{Fraction of high  commitment tweets} (label-3)
We compare the fraction of entities' tweets classified as high commitment to find entities who have large fraction of high commitment tweets but get low action-ratings. This method aims to distinguish between cases: entity A has 10 high-commitment tweets out of 100 total tweets, while entity B has 10 high commitment tweets out of 20 total tweets.

\textbf{Prediction probability of high cause commitment tweets} (label 3)
As we know, higher prediction probability gives more confident of the predicted label. If entity A and B have the same number of predicted label-3 instances, but the prediction probability of label-3 in A is greater than 0.8 while it is around 0.7 in B, then we think A should get higher words-rating. Compare this average prediction probability with entities' action-ratings help us find out entities who have high prediction probabilities but low action-ratings.
}

%% file: data.tex
In this section, we describe our experimental datasets used in the proposed approach\cut{ used to evaluate the proposed approach}. \cut{\textbf{Replication code and data will be released upon publication.}\cut{\footnote{Replication code and data will be released upon publication.}}}

\begin{table*}[t]
\begin{center}
\begin{tabular}{| c | c | c | c || c | c | c | c |}
\hline
 \multicolumn{4}{|c ||} {} & \multicolumn{4}{c |}{\textbf{Ratio of 4 commitment levels}} \\ 
\hline
\textbf{Cause} & \textbf{Entity} & \textbf{Public Message} & \textbf{Labeled Instances} & \textbf{0} & \textbf{1} & \textbf{2} & \textbf{3} \\
\hline
Health & 142 Brands &  429,009 tweets & 426 & 0.023 & 0.241 & 0.381 & 0.355 \\
\hline
Eco & 966 Brands & 2,624,800 tweets & 966 & 0.447 & 0.234 & 0.165 & 0.154 \\
\hline
Eco & 514 Congress members & 1,118,962 tweets & 514 & 0.063 & 0.197 & 0.467 & 0.273\\
\hline
\end{tabular}
\end{center}
\caption{\label{tab.tweets} Twitter data collected for 3 pairs of cause-entity types and corresponding commitment distributions}
\end{table*}

\subsection{Third-party ratings}
We collect entities along with their third-party ratings of environmental actions (for brands and politicians) and health actions (for brands) from the following sources.

\subsubsection{GoodGuide}
GoodGuide\footnote{http://www.goodguide.com/} is a website that provides ratings for products from health aspects, company-level environmental and social issues. Products are scored from 0 to 10.\cut{, where higher rating indicates better health or environmental perspective. The rating system is developed using methodologies grounded in the science of health informatics and environmental risk assessment.} We collect scores for 966 brands across 10 sectors.

\subsubsection{The League of Conservation Voters (LCV)}
LCV's National Environment Scorecard\footnote{https://www.lcv.org/} has provided objective and factual information about environmental legislation (e.g., \textit{global warming, wildlife conservation, and so on}) and has become the standard bearer to determine the environmental record of Congress members since 1970.
We collect scores for 514 Congress members.

\subsection{Twitter}
We choose Twitter as the public messaging platform. For each of the entities collected above(\textit{brands and congress members}), we identify the corresponding Twitter account, and download the most recent 3,200 tweets from each account. Table~\ref{tab.tweets} shows details.

\cut{In marketing domain, the collected 966 brands are distributed in 10 sectors (e.g., food, personal care, electronics, and so on). For the environmental issue, we explore all of 966 brands, but only consider 142 brands from Food and Personal Care sectors for the health issue (since other sectors have no ratings with respect to health). Table~\ref{tab.tweets} provides details.}

\cut{
\begin{table}[h]
\begin{center}
\begin{tabular}{| p{1cm} | p{1.5cm} | p{1.5cm} | p{1.5cm} |}
\hline
\multicolumn{1}{|c|}{\textbf{Cause}} & \multicolumn{1}{|c|}{\textbf{Entity}} & \multicolumn{1}{|c|}{\textbf{Public Message}} & \multicolumn{1}{|c|}{\textbf{Labeled Instances}} & 
\multicolumn{1}{|c|}{\textbf{Ratio of 4 commitment levels}} \\
\hline
Health & 142 Brands &  429,009 tweets & 426 \\
\hline
Eco & 966 Brands & 2,624,800 tweets & 966 \\
\hline
Eco & 514 Congress members & 1,118,962 tweets & 514 \\
\hline
\end{tabular}
\end{center}
\caption{\label{tab.tweets} Twitter data collected for 3 entity & causes pairs}
\end{table}
}
\cut{
\subsubsection{Training instances selection}
Using the annotation criteria from Table~\ref{tab.criteria}, we manually annotated a sample of tweets for each task. To do so, we selected the top tweet from each account based on the topic relevance scoring metric described in Section~\ref{sec.topic}. Table~\ref{tab.tweets} provides detailed statistics.}

%% file: results.tex
We use datasets in Table~\ref{tab.tweets} to validate the proposed approach\footnote{Code is available here: https://github.com/tapilab/icdm-2017-causes} (\textit{Note: All manually labeling-processes reached 90\% agreement among annotators}). We focus on 5 research questions:
\begin{itemize}
    \item {\noindent{\bf Sentiment:}} Do high commitment tweets express more positive sentiment towards a cause?\cut{What is the sentiment distribution in each label? Does sentiment imply commitment level?}
    \item {\noindent {\bf Support classification:}} How well can we distinguish between non-support (\textit{labels: 0,1}) and support (\textit{labels: 2,3}) classes? This indicates which tweets express at least a weak support for a cause.
    \item {\noindent {\bf Commitment classification:}} How well can we distinguish between low-commitment (\textit{label-2}) and high-commitment (\textit{label-3}) classes?\cut{ This allows us to differentiate between low and high commitment to a cause.}
    \item {\noindent {\bf Correlation with third-party ratings:}} Do entities that tweet more about supporting a cause actually have high ratings with respect to that cause? Does distinguishing between low- and high- commitment tweets provide a stronger signal of third-party ratings?
    \item {\noindent {\bf Inauthenticity detection:}} If an entity has a low third-party rating but shows high commitment in tweets, is this indicative of possibly inauthentic public messaging?
\end{itemize}

\subsection{Relationship between sentiment and commitment}
\label{sec:5.1}
We use the TextBlob\footnote{https://textblob.readthedocs.io/en/dev/} Python library to classify labeled tweets.\cut{In order to investigate the relationship between a message's sentiment and commitment level, we use the TextBlob\footnote{https://textblob.readthedocs.io/en/dev/} Python library to classify the labeled tweets by sentiment, which in turn uses the Pattern library\footnote{http://www.clips.ua.ac.be/pages/pattern} to classify based on lexical matching.} Table~\ref{tab.sentiment} shows ratios of positive, negative and neutral tweets in 4 commitment labels of 3 datasets. Across all datasets, most instances have positive or neutral sentiment and only few have negative sentiment. However, there does not appear to be a strong relationship between positive sentiment and commitment level, which means positive sentiment alone is not sufficient to distinguish between different commitment levels, as suggested by our discussion in Section~\ref{sec:relatedwork}.\cut{ For example, in the brand-eco data, 54.4\% of label-2 tweets(low-commitment) and 45.3\% of label-3 tweets(high-commitment) are labeled with positive sentiment. Thus, it appears that positive sentiment alone is not sufficient to distinguish between levels of commitment, as suggested by our discussion in Section~\ref{sec:relatedwork}.}

\begin{table}[t]
\begin{center}
\begin{tabular}{| p{0.4cm} || p{0.4cm} | p{0.4cm} | p{0.4cm} || p{0.4cm} | p{0.4cm} | p{0.4cm} || p{0.4cm} | p{0.4cm} | p{0.4cm} |}
\hline
  \multicolumn{1}{|c||}{\textbf{Label}}
  & \multicolumn{3}{c ||}{\textbf{Brand Health}}  & \multicolumn{3}{c ||}{\textbf{Brand Eco }} & \multicolumn{3}{c |}{\textbf{Congress Eco }} \\ 
\hline
 & \textbf{pos} & \textbf{neg} & \textbf{neu} & \textbf{pos} & \textbf{neg}  & \textbf{neu} & \textbf{pos} & \textbf{neg} & \textbf{neu} \\
\hline
\textbf{0} & 0.60 & 0.00 & 0.40 & 0.61 & 0.09 & 0.30 & 0.56 & 0.10 &0.34 \\
\hline
\textbf{1} & 0.70 & 0.05 & 0.25 & 0.55 & 0.12 &0.33 & 0.55 & 0.16 & 0.29\\
\hline
\textbf{2} & 0.82 & 0.05 & 0.13 & 0.54 & 0.09 & 0.37 & 0.54 & 0.09 & 0.37\\
\hline
\textbf{3} & 0.79 & 0.06 & 0.15 & 0.45 & 0.10 &0.45 & 0.45 & 0.09 & 0.46\\
\hline
\end{tabular}
\end{center}
\caption{\label{tab.sentiment} Ratio of sentiments in 4 commitment labels for 3 datasets}
\end{table}

\cut{
\begin{figure}[t]
\centering
\subfloat{
	\includegraphics[width=0.45\textwidth]{fig/brand_health11.png} } 
 
\subfloat{
	\includegraphics[width=0.45\textwidth]{fig/brand_eco11.png} } 
 
\subfloat{
	\includegraphics[width=0.45\textwidth]{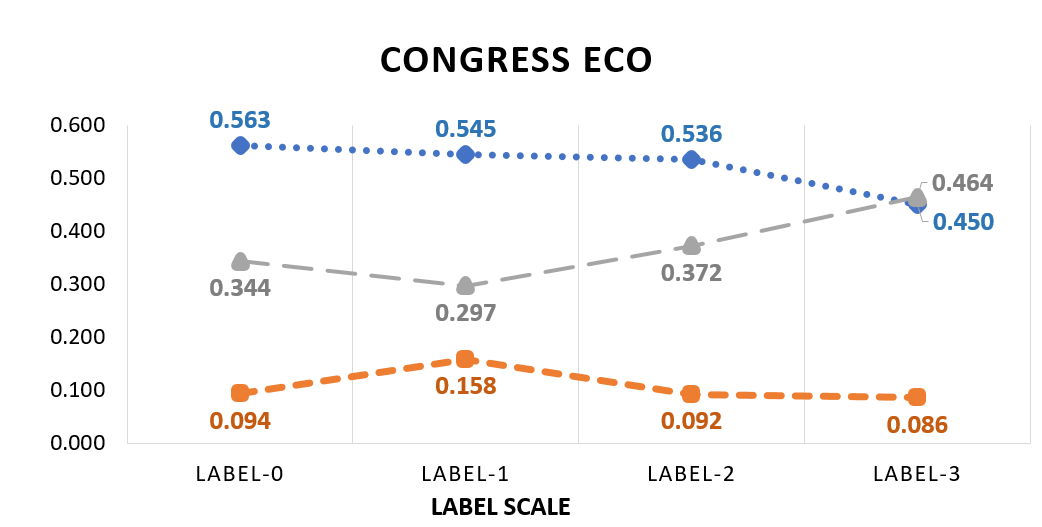}} 
 
\caption{Ratio of positive and negative sentiment in 4 commitment labels for 3 pairs of entity-cause datasets: brand-eco, brand-health, congress-eco}
\label{fig:sentiment}
\end{figure}
}

\cut{Furthermore, in the brand-health dataset, the ratios of positive and negative sentiment are both increasing with labels increasing. In brand-eco and congress-eco dataset, ratios of positive sentiment are decreasing when labels increase, but increasing for neutral sentiment. Thus we conclude: sentiment distributions are not changing with commitment labels in regular pattern, and no evidence shows sentiment analysis is necessary for determining commitment levels in 3 datasets.}

\subsection{Evaluation for support classification }
\label{sec:5.2}
In support classification, we distinguish between non-support class (\textit{labels: 0,1}) and support class (\textit{labels: 2,3}). For each of the 3 datasets, we experiment with a number of different classifiers (e.g., \textit{LogisticRegression, SVM, MLP, DecisionTree}) and use GridSearchCV in scikit-learn\footnote{http://scikit-learn.org} to select the best combination of features and model parameters that give highest F1 score for support class. In this task, LogisticRegression classifier outperforms others.

\begin{figure}[t]
	\centering
	\includegraphics[width = 0.5\textwidth]{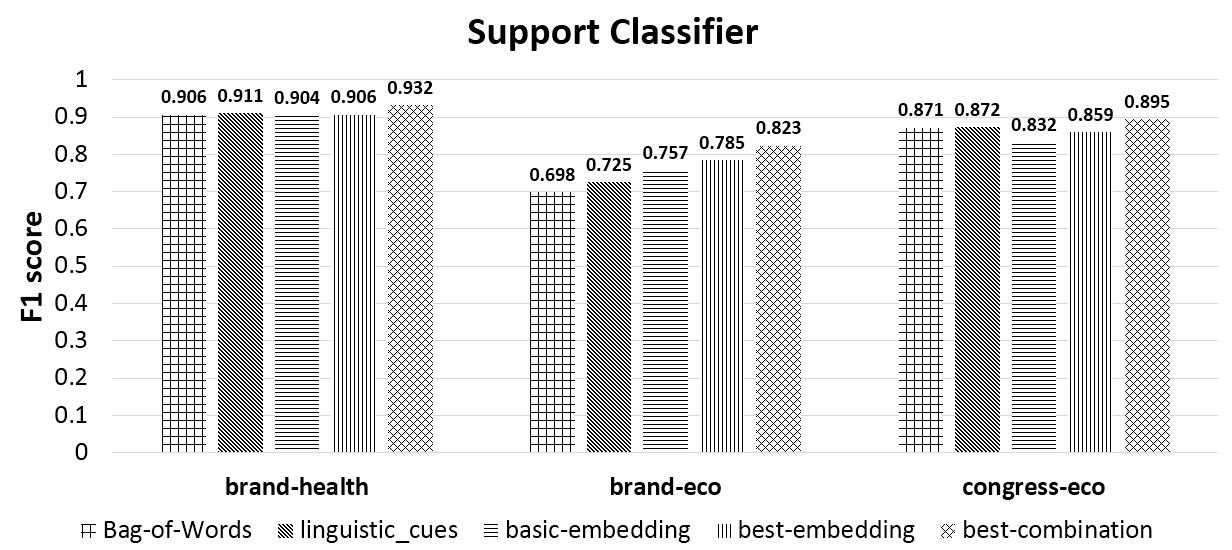}
	\caption{Average 10-fold cross-validation F1 scores of support classifier with different sets of features. Bag-of-Words features serve as baseline, and linguistic-cues refer to features in~\ref{sec:3.3}. Basic-embedding feature refers to tweet vector, and best-embedding refers to the set of embedding features that produce best F1 score. Best-combination is the combination of linguistic features and embedding features that produce best F1 score. }
	\label{fig:sup_cls}
\end{figure}

Fig.~\ref{fig:sup_cls} shows F1 scores of support classifier for 3 datasets and score variations with different features. Meanwhile, Table~\ref{tab.acc} lists details of precision, recall and F1 score for the best classifier. Several conclusions can be drawn from Fig.~\ref{fig:sup_cls}: (1) Performance of Bag-of-Words feature is improved after adding linguistic cues (e.g., \textit{re-tweet: ``RT", mention: ``@", hashtag: ``\#"},\cut{ mark for pronouns: ``first\_person".}); (2) Embedding features alone do not perform better than linguistic cues (embedding features only capture semantics for normal words but not specific symbols (e.g.,``@'') ); (3) Combination of linguistic features and embedding features gives the best F1 score. Linguistic features are effective only for seen words (words in the training set), and embedding features serve as a complementary to generalize to unseen words when they appear in similar context (words appearing in similar contexts have similar meanings and vector representations).

\begin{table}[t]
\begin{center}
\begin{tabular}{| p{0.85cm} | p{0.6cm} || p{0.6cm} | p{0.6cm} | p{0.6cm} || p{0.6cm} | p{0.6cm} | p{0.6cm} |}
\hline
 & & \multicolumn{3}{c ||}{\textbf{support} ($\{0,1\}$vs.$\{2,3\}$)}  & \multicolumn{3}{c |}{\textbf{commitment} ($2$vs.$3$)} \\ 
\hline
\textbf{Entity} & \textbf{Cause} & \textbf{Prec} & \textbf{Rec} & \textbf{F1} &
\textbf{Prec} & \textbf{Rec} & \textbf{F1} \\
\hline
\multirow{1}{*}{Brands} & \multirow{1}{*}{Health} & 0.935 & 0.929 & 0.932 & 0.782 & 0.765 & 0.773\\
\hline
\multirow{1}{*}{Brands} & \multirow{1}{*}{Eco} & 0.860 & 0.789 & 0.823  & 0.800 & 0.714 & 0.755 \\
\hline
\multirow{1}{*}{Congress} & \multirow{1}{*}{Eco} & 0.890 & 0.902 & 0.895 & 0.708 & 0.721 & 0.712\\
\hline
\end{tabular}
\end{center}
\caption{\label{tab.acc} Precision, recall and F1 scores for the best cross-validation results of support and commitment classifiers}
\end{table}

\begin{table*}[t]
\begin{center}
\begin{tabular}{| p{1cm} | p{1cm} | p{3cm} | p{3.8cm} | p{3.1cm} |  p{3.5cm} | }
\hline

\multicolumn{1}{|c|}{\textbf{Entity}} & \multicolumn{1}{c|}{\textbf{Cause}} & 
\multicolumn{1}{c|}{\textbf{No Support} \textit{(labels: 0,1)}} & 
\multicolumn{1}{c|}{\textbf{Support} \textit{(labels: 2,3)}} & 
\multicolumn{1}{c|}{\textbf{Low commit} \textit{(label-2)}} & 
\multicolumn{1}{c|}{\textbf{High commit} \textit{(label-3)}}\\

\hline
Brands & Health & 
\parbox[t]{3cm}{flavor, cheese, sleek, \\animals, chocolate} &
\parbox[t]{3.8cm}{healthy, nutritious, organic,\\ \#vegan, natural} & 
\parbox[t]{3.1cm}{foods,  eat, recipes, diet, \\veggies } & 
\parbox[t]{3.5cm}{our, natural, \#organic,\\ \#nongmo, certified}\\
\hline
Brands & Eco &  
\parbox[t]{3cm}{skin, food, diet, fish, \\natural} & 
\parbox[t]{3.8cm}{sustainable, environment, \\planet, endangered, sustainability } & 
\parbox[t]{3.1cm}{planet, day, great, can, \\second\_person} & 
\parbox[t]{3.5cm}{protect, \_self\_first\_\_person,\\ first\_person, \#sustainable, we }\\
\hline
Congress & Eco &
\parbox[t]{3cm}{lives, rural, industry, jobs, economic}  &
\parbox[t]{3.8cm}{protect, habitats, conservation, epa, forests} &
\parbox[t]{3.1cm}{plant, epa, historic,\\pollution, global} &
\parbox[t]{3.5cm}{I, my, voted, must, \_bill\_ }\\
\hline
\end{tabular}
\end{center}
\caption{\label{tab.coefs} Features that have high coefficients in support and commitment classifiers }
\end{table*}

Table~\ref{tab.coefs} shows features that have high coefficients in support classifier.\cut{(chosen from the top 10 features for each class).} For support classifier, cause keywords play an important role to distinguish between support and non-support classes (e.g., tweets that support \textit{health} tend to use \textit{health keywords such as: natural, healthy, organic}, but tweets don't support health talks more about \textit{flavor and non-healthy aspects: chocolate, cheese, animals' meat and so on}).

\subsection{Evaluation for commitment classification}
\label{sec:5.3}
\begin{figure}[t]
	\centering
	\includegraphics[width = 0.5\textwidth]{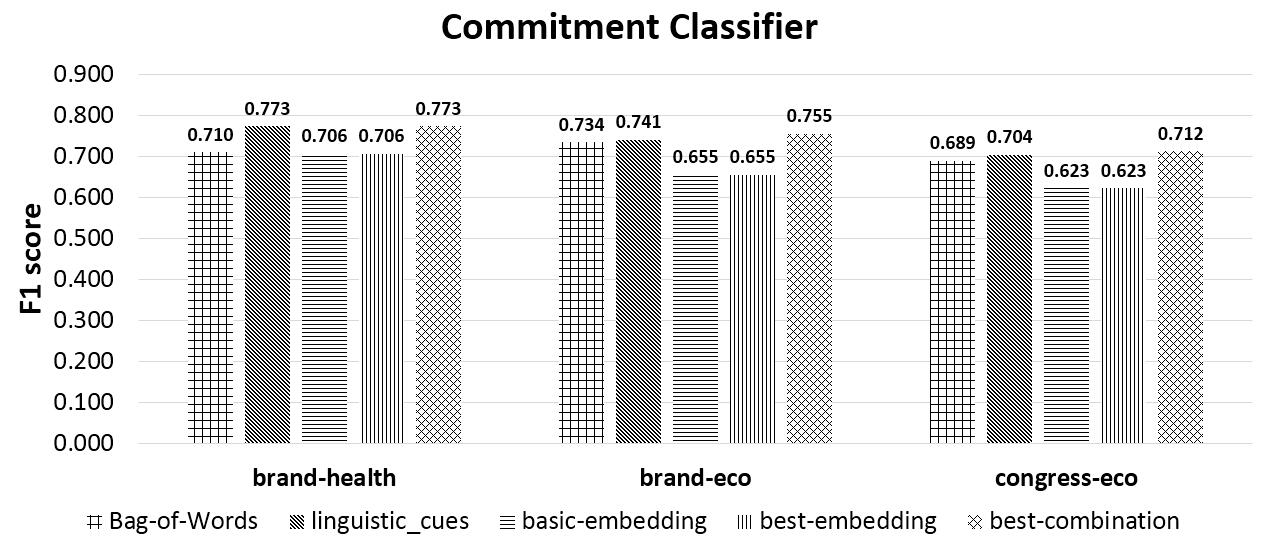}
	\caption{Performance of commitment classifier with different sets of features, please refer to Fig~\ref{fig:sup_cls} for explanation of 5 feature sets.}
	\label{fig:com_cls}
\end{figure}

After classifying tweets into support and non-support classes, we continue to classify support tweets into low- and high- commitment classes with same tuning and evaluation process in support classifier, but the set of features and model parameters that produce best F1 scores are different from support classification. Fig.~\ref{fig:com_cls} shows F1 scores of commitment classifier for 3 datasets and score variations with different features. Table~\ref{tab.acc} lists details of precision, recall and F1 scores for the best classifier.

According to Fig.~\ref{fig:com_cls}, we find that: (1) F1 scores of commitment classifier are lower overall than support classifier, as expected given more nuanced distinction between low- and high- commitment levels; (2) Performance of Bag-of-Words feature is improved again after adding linguistic cues; (3) Embedding features alone perform worse than lexical features due to the lack of vector representations for special and distinguishable characters (e.g., \textit{re-tweet: ``RT", mention: ``@", hashtag: ``\#"}) and thus not capturing subtle differences between low- and high- commitment classes; (4) Combination of linguistic features and embedding features also make improvements (from 0.710 to 0.773 for brand-health, 0.734 to 0.755 for brand-eco,  and 0.689 to 0.712 for congress-eco).

\cut{
\begin{table*}[t]
\begin{center}
\begin{tabular}{| c | c | c | c | c | c | }
\hline
\multirow{2}{*}{\textbf{Entity}} & \multirow{2}{*}{\textbf{Cause}} & \textbf{No Support} & \textbf{Support} & \textbf{Low commit} & \textbf{High commit}\\
 & & \textit{labels $\{0,1\}$} & \textit{labels $\{2,3\}$} & \textit{label $2$} & \textit{label $3$} \\
\hline
Brands & Health & 
\parbox[t]{2.5cm}{flavor, cheese \\ sleek, animals \\ chocolate} &
\parbox[t]{2.5cm}{healthy, nutritious \\ organic, \#vegan \\ natural} & 
\parbox[t]{2.5cm}{foods,  eat \\ recipes, diet \\ veggies } & 
\parbox[t]{2.5cm}{our, natural \\ \#organic, \#nongmo \\ certified}\\
\hline
Brands & Eco &  
\parbox[t]{2.5cm}{skin, food \\ diet, fish \\ natural} & 
\parbox[t]{2.5cm}{sustainable, planet \\ environment,  endangered \\ sustainability } & 
\parbox[t]{2.5cm}{planet, day \\ great, can \\ second\_person} & 
\parbox[t]{2.5cm}{we, protect \\ first\_person, \#sustainable \\ \_self\_first\_\_person }\\
\hline
Congress & Eco &
\parbox[t]{2.5cm}{lives, rural \\jobs, \#energy \\ dimension\_15}  &
\parbox[t]{2.5cm}{protect, habitats \\ conservation, epa \\ dimension\_18} &
\parbox[t]{2.5cm}{rt, you \\ epa, historic \\ \_@self\_pollution} &
\parbox[t]{2.5cm}{I, my \\ voted, must \\ \_bill\_ }\\
\hline
\end{tabular}
\end{center}
\caption{\label{tab.coefs} Features that have high coefficients in support and commitment classifiers }
\end{table*}
}

Table~\ref{tab.coefs} shows features that have high coefficients in commitment classifier for each dataset.\cut{(chosen from the top 10 features for each class)}\cut{It shows that: For commitment classifier} We find: the use of imperative words (e.g., \textit{let, must, need}), pronouns (\textit{I, me ,we}), @oneself together with action verbs (\textit{vote, conserve}) provide the strongest signals (These features corresponds to definition of high-commitment in table~\ref{tab.criteria}). Among these distinguishable features, most of them are discarded when generating embedding features (e.g., pre-trained GoogleNews Word2Vec model has no vector representation for @oneself), which explains the bad performance of embedding features in Fig.~\ref{fig:com_cls}.

\cut{
\begin{table}[t]
\begin{center}
\begin{tabular}{| c | c | p{2.5cm} | p{2.5cm} | }
\hline
\multirow{2}{*}{\textbf{Entity}} & \multirow{2}{*}{\textbf{Cause}} & \textbf{Low commit} & \textbf{High commit}\\
 & &  label $2$ & label $3$ \\
\hline
Brands & Health & 
\parbox[t]{2.5cm}{foods, eat, recipes, veggies, diet\\} & 
\parbox[t]{2.5cm}{our, \#organic, natural, \#nongmo, certified\\}\\
\hline
Brands & Eco &  
\parbox[t]{2.5cm}{planet, day, great, can, second\_person\\} & 
\parbox[t]{2.5cm}{we, first\_person,  \_self\_first\_\_person, protect, \#sustainable}\\
\hline
Congress & Eco &
\parbox[t]{2.5cm}{rt, you, epa, historic, \_self\_pollution} &
\parbox[t]{2.5cm}{I, my, voted, \_bill\_  must}\\
\hline
\end{tabular}
\end{center}
\caption{\label{tab.com_coefs} Top coefficient features in commitment classifiers }
\end{table}
}

\subsection{Correlation with third party ratings}
\label{sec:5.4}

We conduct a regression analysis to quantify how the volume of non-support, low- and high- commitment tweets correlate with third-party ratings. In this analysis, to reduce prediction noise, we only take instances with prediction probability greater than 0.7 as confident predictions (though results are similar without this threshold). 
We fit an Ordinary Least Squares regression model as follows:
\begin{center}
$\mathtt{y}=\beta_0 + \beta_1*\mathtt{x_1}+\beta_2*\mathtt{x_2}+\beta_3*\mathtt{x_3}$
\end{center}

$\mathtt{y}$ refers to third party ratings, $\mathtt{x_1}$, $\mathtt{x_2}$, $\mathtt{x_3}$ represent log of non-support (\textit{labels: 0, 1}), low-commitment (\textit{label-2}), high-commitment (\textit{label-3}) class frequencies respectively. $\beta_1$, $\beta_2$ and $\beta_3$ are corresponding coefficients.

\begin{figure}[t]
	\centering
	\includegraphics[width = 0.5\textwidth]{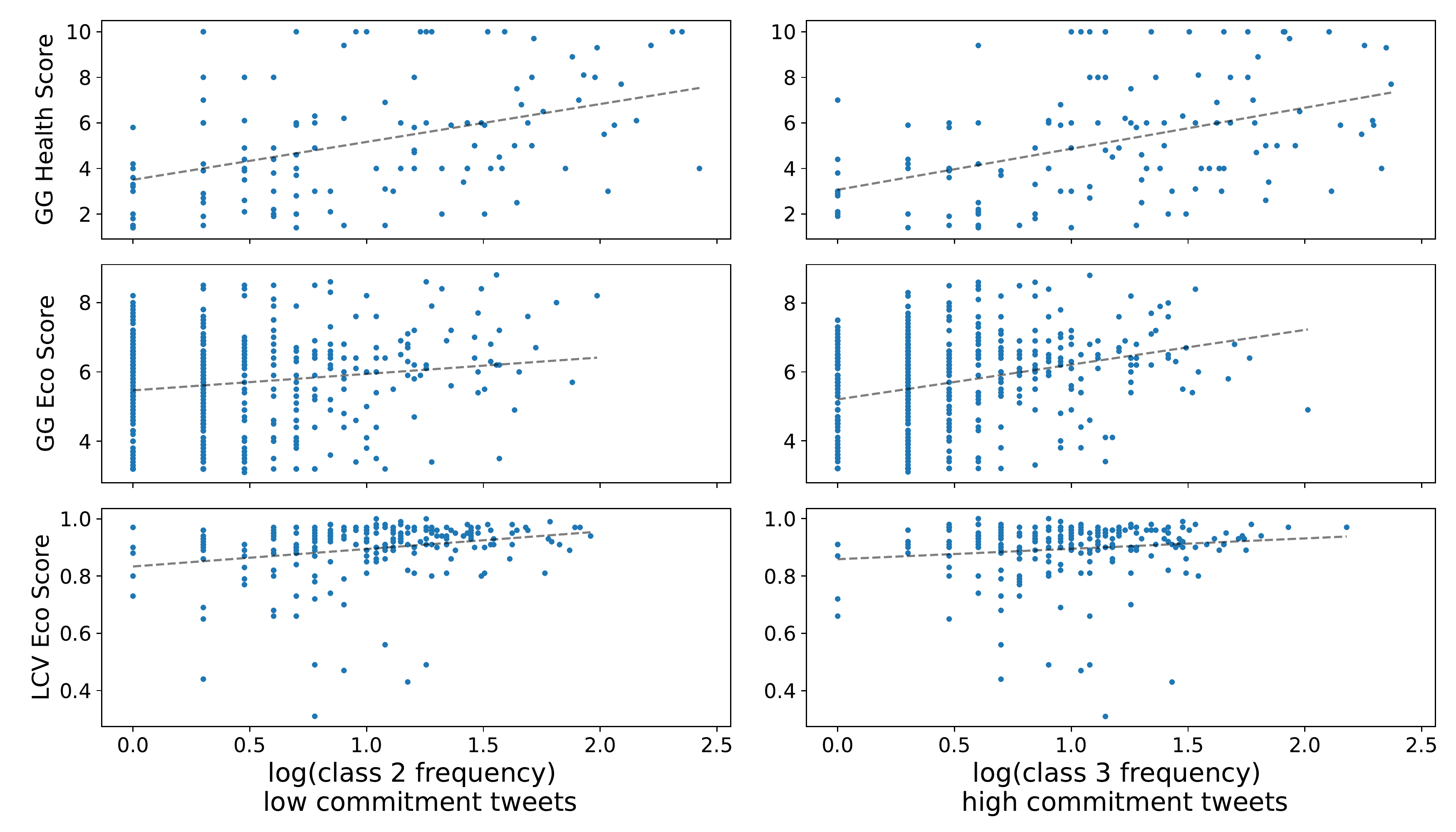}
	\caption{Scatter plots for coefficients of low- and high- commitment tweets with third-party ratings}
	\label{fig:scatter}
\end{figure}

Fig.~\ref{fig:scatter} shows scatter plots of how the volume of low- and high- commitment classes independently correlate with third party ratings. And Table~\ref{tab.coef} lists the coefficients and p-values for each class in 3 datasets (\textit{GG refers to GoodGuide}).

\begin{table}[t]
\begin{center}
\begin{tabular}{| p{1.3cm} || p{0.75cm} | p{0.7cm} || p{0.73cm} | p{0.62cm} || p{0.73cm} | p{0.62cm} |}
\hline
 & \multicolumn{2}{c ||}{\textbf{GG HealthScore}}  & \multicolumn{2}{c ||}{\textbf{GG EcoScore }} & \multicolumn{2}{c |}{\textbf{LCV EcoScore }} \\ 
\hline
 & \textbf{coef} & \textbf{p-val} & \textbf{coef} & \textbf{p-val} & \textbf{coef} & \textbf{p-val}\\
\hline
Non-sup & -1.501 & 0.033 & -0.206 & 0.257 & -0.034 & 0.202 \\
\hline
Low-comt & 1.168 & 0.015 & 0.101 & 0.614 & 0.027 & 0.149 \\
\hline
High-comt & 1.451 & 0.004 & 1.092 & 0.001 & 0.029 & 0.066 \\
\hline
\end{tabular}
\end{center}
\caption{\label{tab.coef} coefficients and p-values for non-support, low-/high- commit classes' log frequencies with respect to third-party ratings}
\end{table}

Note that we do not expect a very strong correlation, given the expected presence of inauthentic messaging. Additionally, there are many outliers for small values of the x-axis, particularly for the GG eco score. This arises because there are many brands with high third-party eco ratings that nonetheless do not have many tweets identified as high-commitment, which suggests that some firms may not view environmental marketing as valuable, either because of the risk of being perceived as inauthentic, or because marketing managers do not believe such messaging fits into the overall personality of the brand. These results also explain why it is not advisable to fit a text classification model directly to the third party ratings -- given the misalignment between words and actions, such a classifier would likely overfit to the outliers in these data.

Besides that, Table~\ref{tab.coef} shows: (1) For all datasets, non-support classes have negative coefficients with third party ratings while low- and high- commitment classes have positive coefficients with third-party ratings, which means \textbf{entities that tweet more about supporting a cause actually have high third-party ratings}; (2) For all datasets, high-commitment classes show higher coefficients than low-commitment classes, suggesting that \textbf{high-commitment classes provide stronger signal of third-party ratings than low-commitment classes}. 
\cut{
We may interpret the correlation between high-commitment tweets and third-party rating as an approximate measure of the alignment between words and actions in a domain. It is notable, then, that the strongest relationship is in the brand-health data\cut{ (as indicated by the magnitude of the standardized coefficient for the high-commitment class variable)}, followed by brand-eco and congress-eco.\cut{ This makes some intuitive sense, as the health claims of a food product are more highly regulated than claims of environmental friendliness; furthermore, given that nutrition is a more immediately observable property of a product than environmental friendliness, consumers may be more sensitive to inauthentic messages in this domain. To more rigorously make such cross-domain comparisons will require additional data sources, which we leave for future work.}
}

In next section, we will investigate more closely outliers in scatter plots that may be indicative of inauthentic messaging.

\subsection{Detecting potentially inauthentic entities}
\label{sec:5.5}
As described in Section \ref{sec:aggregation}, we first select a set of entities who show high-commitment in public messaging and then mark those who have low third-party ratings as potentially ``inauthentic" entities. Below, we provide both quantitative and qualitative analysis \cut{of the results}to assess the feasibility of this approach. We emphasize that the identification of these entities does not necessarily indicate any wrong doing --- there may be valid reasons for the\cut{discovered} misalignment, which we explore below. 
\subsubsection{Quantitative analysis}
\label{sec:5.5.1}
For a quantitative measure, for each of the 3 datasets, we select 10 detected ``inauthentic" entities along with their tweets that are predicted as high-commitment with probability greater than 0.7.
\cut{
(the same threshold used to identify them)
}We then manually annotate these tweets to compute the precision of the high-commitment classifier. This provides both an additional validation measure for the classifier on unseen data, as well as a check to ensure that the tweets have been correctly identified as high-commitment.

\begin{table}[t]
\begin{center}
\begin{tabular}{| c |  c | c | }
\hline
\thead{\textbf{Entity}} & \thead{\textbf{Cause}} & \thead{\textbf{Precision}}\\
\hline
Brand & Health & 73.74\% \\
\hline
Brand & Eco & 94.80\% \\
\hline
Congress & Eco & 75.93\% \\
\hline
\end{tabular}
\end{center}
\caption{\label{tab.ratio} Precision of tweets (\textit{from low-rated entities}) that are classified as high-commitment}
\end{table}

Table~\ref{tab.ratio} shows the percentage of correctly predicted high-commitment tweets for each domain. Precisions \cut{is about 75\%  }for brand-health and congress-eco datasets are consistent with the performance of commitment classifier. However, we find higher accuracy for the brand-eco dataset. This may in part be because some entities post many similar tweets and if one of them is correctly classified, then all of them are correctly classified. \cut{(e.g.,\textit{``@singalldaylong Hi Emma. Our commitment to health includes environmental health as well."} is similar to \textit{``@RyDizy CVS Pharmacy’s commitment to health includes environmental health as well."}).}
\subsubsection{Qualitative analysis}
\label{sec:5.5.2}
For a qualitative evaluation, we manually read the tweets of each entity to develop a better understanding of each domain. \cut{We will look at each domain below.}

In the congress-eco domain, the top three identified entities are moderate Democratic members from swing districts (Rep. Ann Kirkpatrick from Arizona, Rep. Kurt Schrader from Oregon and Rep. Joe Manchin from West Virginia). These members often tweet more narrowly about conservation and wildlife protection, as opposed to more broad appeals to prevent global climate change, which is common among more liberal Congress members. Rep. Kirkpatrick was a Democratic member of the House of Representatives until January 2017, representing Arizona's 1st congressional district, which is known to be a swing district --- it voted for Republican presidential candidates in the last 5 elections, and of the most recent 6 representatives, 3 were Republicans, and 3 were Democrats. Thus, we would expect a politician in such a district to express more nuanced views toward the environment to cater to such a heterogeneous constituency~\cite{jacobson2015politics}. Indeed, we find that Rep. Kirkpatrick has a very low lifetime environmental rating from the LCV (68\%), which is the 7th lowest score given to a Democratic member of the House in the 2016. Despite such a low score, Rep. Kirkpatrick had 11 tweets classified as high commitment, of which 8 were manually verified as high commitment. For example, one message was a retweet from a non-profit in Arizona that focuses on energy and conservation (``{\it RT @SonoranArizona: Great meeting with \@RepKirkpatrick staff on I-11, renewable energy and conservation...}"). Many of the other messages involve her work on a bill to prevent forest fires in Arizona, which was often framed as conservation (``{\it Thx to \@CGDispatch for covering passage of my bipartisan bill to protect \#AZ forests...}"). Thus, it appears that in the politics domain, entities may frame their public messaging to emphasize their actions with respect to a smaller, bipartisan subset of issues within the overall cause.

For brand entities, we find among the identified entities a number of brands whose public messaging attempt to align themselves with exercise and fitness, even though the product may not be so aligned. Table~\ref{tab.inauthentic} lists 3 examples from brands with low health ratings. Little Debbie produces many pre-packaged desserts, such as mini-cakes and brownies, that are high in saturated fat and sugar. Some of their messages reference how these snacks may be helpful for those training for an Iron Man race (\#IM-Chattanooga). Amp Energy is an ``energy drink" produced by PepsiCo. A variant of ``Mountain Dew," the drink is mainly popular because of its high caffeine. \cut{Some of the tweets from this brand also attempt to align themselves with fitness and exercise, even though some flavors of the drink contain as much sugar as a Coca-Cola.}

\begin{table}[t]
\begin{center}
\begin{tabular}{| c | c | p{5.5cm} | }
\hline
\multicolumn{1}{|c|}{\textbf{Brand}} & 
\multicolumn{1}{c|}{\textbf{Score}} & 
\multicolumn{1}{c|}{\textbf{Inauthentic high-commitment tweets}}\\
\hline
littledebbie & 1.5 & \textit{``RT @quintanarootri: Do you have a @LittleDebbie \textbf{nutrition} plan for \#IMChattanooga? Simple \textbf{carbs} for quick \textbf{energy} on the \textbf{bike} \#itspersonal"} \\
\hline
ampenergy & 1.4 & \textit{``\textbf{Exercising} in the morning helps you stay \textbf{energized} throughout the day. Pack your bag the night before to make getting to the \textbf{gym} easier!"} \\
\hline
sprite & 1.5 & \textit{@Randa\_Rocks   Randa, we have \textbf{vitaminwater zero} made with \textbf{stevia}. No other products yet but we’re always coming up with \textbf{new ideas}!} \\
\hline
\end{tabular}
\end{center}
\caption{\label{tab.inauthentic} Examples of tweets classified as high-commitment from entities identified as potentially ``inauthentic"}
\end{table}

As Table~\ref{tab.ratio} suggests, precisions are not 100\%, which means some entities may possibly be identified potentially ``inauthentic" due to mis-classification. For example, the Sprite tweet in Table~\ref{tab.inauthentic} emphasizes that a sugar substitute used in their vitamin water. This may be interpreted as only a weak indicator of support. Overall, however, we find that all identified entities have at least one tweet manually verified as high-commitment.

Taken together, these results suggest that the proposed approach can identify potentially ``inauthentic" entities, but we recommend manual verification\cut{ of the discovered entities and tweets} to understand more precisely how the language relates to the cause being considered.

\cut{
\begin{table*}[t]
\begin{center}
\begin{tabular}{| p{3cm} | p{1cm} | p{10cm} | }
\hline
\multicolumn{1}{|c|}{\textbf{Entity}} & \multicolumn{1}{|c|}{\textbf{Cause}} & \multicolumn{1}{|c|}{\textbf{Detected Potentially Inauthentic Entities}} \\
\hline
Brands & Health & littledebbie, monsterenergysa, cocacola, ampenergy, sprite, welchs, honeymaidsnacks, honeystinger, miessence, gardein, 5hourenergy \\
\hline
Brands & Eco & ecotools, bayer, alaffiaskincare, fritolay, cvs\_extra, maggiesorganics, quakeroatsus \\
\hline
Congress & Eco & RepJimCosta, Sen\_JoeManchin, SenatorHeitkamp, RepVisclosky, RepCuellar, SenatorReid \\
\hline

\end{tabular}
\end{center}
\caption{\label{inauthentic-table} Detected  entities who tweet more about the cause than expected based on their third-party ratings.}
\end{table*}
}

%% file: conclusion.tex
This paper proposes a framework to investigate how entities talk and how they act. We use Twitter public messaging as source of entities'  words, and third-party ratings as a measure of how they act. Entities who post high-commitment tweets but have low third-party ratings are detected as ``inauthentic''\cut{, which means their public messaging does not align with their actions}. 

This framework can be generalized to any domain that has a collection of short texts with corresponding ratings (For example, Amazon's/ebay's product descriptions and ratings, restaurants' menu descriptions and customers' votings, and so on). The proposed commitment labeling criteria (Table~\ref{tab.criteria}) is invariant to different domains. The linguistic features may be a little different for distinct domains but the idea of exploring and combining linguistic cues and embedding vectors to enrich feature representation for short texts can be applied to various domains. \cut{Last but not the least important thing is: we give substantive analysis of the proposed research questions to make the experimental results transparent.}

However, there are several limitations need to address in our future work: (1) Our training datasets are manually labeled by experts, and we will automate this task in future work. (2) Once the labeling process is automated, we can get larger size of training data, and then apply more complex techniques such as deep learning. (3) We use pre-trained GoogleNews Word2Vec model to get word vectors, however, this model has no vector representation for some specific but important features in public messaging (e.g., \textit{re-tweet, self-mention: @one's-name, hash-tags: \#event}), and we will train Word2Vec models for public message to deal with this problem. (4) Also, we will search for more sources of third-party ratings to check for robustness of this framework.

\cut{
We first present a text classification approach to categorize messages according to the commitment level they express towards a cause. For text representation, we explore a number of features, such as word embeddings, polarity, and pronoun usage. Experiments show that adding Word2Vec features significantly increases classification accuracy. Besides, we find several vector dimensions correlate with target cause.

We then apply the supportive classifier and commitment classifier to all the historical tweets of hundreds of entities, and quantify the volume of tweets assigned to each category of commitment. We conduct regression analysis to compare correlation of volume of high- and low- commitment messages with third party ratings and find that high-commitment tweets show stronger correlation with third-party ratings than low-commitment tweets. Based on tweets' commitment levels, we use aggregation methods to select a set of high-commitment entities.

Finally, by measuring the discrepancy between entities' commitment in public messaging and their action-ratings collected from third-party, we identify entities that appear to express a stronger commitment to a cause in their messaging than in their actions as ``inauthentic'' entities. 

Although these basic features and models are efficient in this task, there's still prospecting improvements with further exploration of high-level features and advanced models in future work.
}